\documentclass[lettersize,journal]{IEEEtran}
\usepackage{amsmath,amsfonts}
\usepackage{algorithmic}
\usepackage{algorithm}
\usepackage{array}
\usepackage[caption=false,font=normalsize,labelfont=sf,textfont=sf]{subfig}
\usepackage{textcomp}
\usepackage{stfloats}
\usepackage{url}
\usepackage{verbatim}
\usepackage{graphicx}
\usepackage{cite}
\usepackage{amssymb}
\usepackage{booktabs}
\usepackage{multirow}
\usepackage{bbm}
\usepackage{algorithm}
\usepackage{algorithmic}
\usepackage{caption}
\usepackage[dvipsnames]{xcolor}
\usepackage{colortbl} 
\newcommand\best[1]{\textbf{\color{RedOrange}#1}} 
\newcommand\secondbest[1]{\textbf{\color{Periwinkle}#1}}  
\definecolor{Gray}{gray}{0.85}
\newcolumntype{a}{>{\columncolor{Gray}}c}
\newcommand\ie{\textit{i.e.}}
\newcommand\eg{\textit{e.g.}}

\usepackage[pagebackref,breaklinks,colorlinks]{hyperref}
\usepackage[capitalize]{cleveref} 
\crefname{section}{Sec.}{Secs.} 
\Crefname{section}{Section}{Sections} 
\Crefname{table}{Table}{Tables} 
\crefname{table}{Tab.}{Tabs.} 
\begin{document}

\title{MAPS: A Noise-Robust Progressive Learning Approach for Source-Free Domain Adaptive Keypoint Detection}

\author{Yuhe Ding,
Jian Liang,
Bo Jiang,
Aihua Zheng
and~Ran~He,~\IEEEmembership{Senior~member,~IEEE}
\thanks{Yuhe Ding and Bo Jiang are with the School of Computer Science and Technology, Anhui University. E-mail: madao3c@foxmail.com; jiangbo@ahu.edu.cn.}
\thanks{ Aihua Zheng is with the School of Artificial Intelligence, Anhui University. E-mail: ahzheng214@foxmail.com.}
\thanks{Jian Liang and Ran He are with the National Laboratory of Pattern Recognition (NLPR) and the Center for Research on Intelligent Perception and Computing, CASIA. E-mail: liangjian92@gmail.com; rhe@nlpr.ia.ac.cn.}
\thanks{Jian Liang is the corresponding author.}
}
\markboth{Journal of \LaTeX\ Class Files,~Vol.~14, No.~8, August~2021}%
{Shell \MakeLowercase{\textit{et al.}}: A Sample Article Using IEEEtran.cls for IEEE Journals}


\maketitle

\begin{abstract}
Existing cross-domain keypoint detection methods always require accessing the source data during adaptation, which may violate the data privacy law and pose serious security concerns.
Instead, this paper considers a realistic problem setting called source-free domain adaptive keypoint detection, where only the well-trained source model is provided to the target domain.
For the challenging problem, we first construct a teacher-student learning baseline by stabilizing the predictions under data augmentation and network ensembles.
Built on this, we further propose a unified approach, Mixup Augmentation and Progressive Selection (MAPS), to fully exploit the noisy pseudo labels of unlabeled target data during training.
On the one hand, MAPS regularizes the model to favor simple linear behavior in-between the target samples via self-mixup augmentation, preventing the model from over-fitting to noisy predictions.
On the other hand, MAPS employs the self-paced learning paradigm and progressively selects pseudo-labeled samples from `easy' to `hard' into the training process to reduce noise accumulation.
Results on four keypoint detection datasets show that MAPS outperforms the baseline and achieves comparable or even better results in comparison to previous non-source-free counterparts.
The code will be released at \url{https://github.com/YuheD/MAPS}.
\end{abstract}


\section{Introduction}
Unsupervised domain adaptation (UDA), a paradigm that aims to transfer knowledge from a label-rich source domain to another unlabeled target domain, is proposed to reduce the burden of manual annotations \cite{csurka2017comprehensive}.
Typically, existing UDA methods resort to feature alignment \cite{long2015learning, tzeng2017adversarial} or domain translation \cite{murez2018image} so that the discriminative model learned on the labeled source domain could perform well in the unlabeled target domain.
However, these methods always require access to the source data when adapting to the target domain, which may be infeasible due to the growing restrictions on data privacy and storage.
To tackle this issue, a line of recent works \cite{liang2020we, huang2021model, yang2021exploiting, wang2021give} consider a source data-free domain adaptation (SFDA) scheme in which only the well-trained classification model from the source domain is provided for the target domain.
Besides the widely-studied image classification task \cite{liang2020we}, the SFDA scheme has also been extended to other visual tasks, \eg, semantic segmentation \cite{wang2021give, yang2022source, kundu2022balancing, liu2022unsupervised}, object detection \cite{li2022source, liu2022source, zhang2021source} and point cloud recognition \cite{yang2021exploiting,ding2022proxymix}.
Though much effort has been devoted to the classification problem, only a few works \cite{liu2022sourcereg} explore the SFDA scheme for regression problems.

This paper for the first time studies the SFDA scheme in a popular regression task---image keypoint detection---which aims to localize the predefined object keypoints in 2D images \cite{cao2017realtime}.
Prior domain adaptive keypoint detection works typically align the source and target domains by elaborately designed alignment strategies, \eg, domain adversarial training \cite{li2021synthetic,jiang2021regressive} and style transfer model \cite{kim2022unified}.
However, all of them require the existence of source data and target data at the same time.
Besides, current SFDA methods for classification always enlarge the margins of boundaries between different classes on the target domain, \eg, pseudo-labeling \cite{yang2021exploiting, liang2020we}, information maximization \cite{liang2020we}, contrastive learning \cite{huang2021model}.
Such techniques are also not applicable to the keypoint detection task since the regression space is continuous without a clear decision boundary.

For the challenging SFDA keypoint detection task, we come up with a simple baseline method by exploiting the mean-teacher framework \cite{tarvainen2017mean}, where the teacher model is an exponential moving average (EMA) of the student network.
To ensure consistency across models, target samples with different augmentations are utilized in the teacher and student models, and the mean squared error between their predictions is minimized.
This method only fine-tunes the model with the consistency loss and no longer requires access to the source data.
Nevertheless, it does not fully explore the semantic information in the unlabeled target domain, leading to sub-optimal adaptation performance. 
Therefore, we resort to pseudo-labeling \cite{lee2013pseudo}, a popular semi-supervised learning technique that generates the pseudo labels for unlabeled data. 
Since there is no labeled data, solely relying on these noisy pseudo labels is inevitably harmful due to noise accumulation, misleading the model over time. 

In this paper, we propose a new approach termed Mixup Augmentation and Progressive Selection (MAPS) by alleviating the noise accumulation issue for source-free keypoint detection. 
Firstly, MAPS mixes the target samples with their shuffle version to construct the self-mixup augmentation \cite{zhang2017mixup}.
This manner regularizes the model to produce outputs that vary linearly with the inputs, preventing over-fitting to outliers and thereby improving the model's robustness.
Secondly, to explore reliable pseudo labels, MAPS adopts the self-paced learning scheme (SPL) \cite{kumar2010self} to alternately select easy samples with high confidence and apply a standard regression loss on them.
In fact, the number of selected samples is governed by a dynamic weight that is annealed until the training loss converges.
Prioritizing learning about easy samples is proven to prevent the model from getting stuck in a bad local optimum \cite{basu2013teaching}.
These two strategies respectively consider the convex behavior among samples and the quality of the pseudo label per sample, making them complement each other.
To validate the effectiveness of our method, we conduct three experiments on human, hand, and animal keypoint detection.
The main contributions of our work are summarized as follows:
\begin{itemize}
  \item We formulate a realistic and challenging task, \emph{source-free domain adaptive keypoint detection}, which is among the first to study cross-domain regression problems without the source data.
  \item Built on the devised baseline via mean-teacher, we propose a new approach termed Mixup Augmentation and Progressive Selection (MAPS) which fully exploits the noisy pseudo labels in the target domain.
  \item Experiments on various keypoint detection benchmarks demonstrate that MAPS outperforms the baseline and achieves results comparable, or even better than, existing non-source-free counterparts.
\end{itemize}

\section{Related Work}
\subsection{Source-free Domain Adaptation}
As a typical problem in transfer learning \cite{pan2009survey, csurka2017comprehensive, toldo2020unsupervised}, unsupervised domain adaptation (UDA) \cite{csurka2017comprehensive, long2015learning, tzeng2017adversarial, saito2018maximum} aims to improve the performance of the model on an unlabeled target domain with the aid of label-rich source domain.
However, as data privacy has received increasing attention in recent years, source-free domain adaptation (SFDA) \cite{liang2020we} is proposed to solve the UDA problem with only a well-trained source model.

There are two primary types of SFDA approaches, \ie, generation-based, and self-training-based. 
Generation-based methods \cite{li2020model, tian2021vdm, qiu2021source, du2021generation, yao2021source} attempt to restore the source features or images to compensate for the absence of source data. 
Self-training-based methods \cite{liang2020we, huang2021model, ding2022proxymix, ahmed2022cross, ding2022source} leverage self-supervised techniques to explore the intrinsic structure of the target domain.
In recent years, there are also some practices \cite{yan2021source,du2021generation} to solve this problem by selecting part of the target data as a pseudo source domain, to compensate for the unseen source domain.
The following works develop kinds of variants of SFDA, \eg, active SFDA \cite{wang2022active}, black-box SFDA \cite{liang2022dine}, multi-source SFDA \cite{shen2022benefits}, semi-supervised SFDA \cite{ma2021semi}.
In reality, the vanilla SFDA along with these variants is mainly applied in multiple visual classification tasks, \eg, object recognition \cite{liang2020we}, semantic segmentation \cite{wang2021give, kundu2022balancing}, and object detection \cite{li2022source}. 
The SFDA scheme for classification always enlarges the margins of decision boundaries between different classes on the target domain, thereby improving the generalization performance. 
However, the regression space is continuous without a clear boundary, leading to these methods cannot apply to regression tasks directly.
There are a few works proposed for measurement shift \cite{eastwood2021source}, blind image quality assessment \cite{liu2022sourcereg}.
For the general keypoint detection task, it is still blank.
To the best of our knowledge, we are the first to investigate the SFDA scheme on the keypoint detection task.

\subsection{Domain Adaptive Keypoint Detection}
Keypoint detection is a fundamental visual problem.
It is laborious and time-consuming to collect the labeled data for traditional keypoint detection methods\cite{li2021human, newell2016stacked, xiao2018simple, duan2019trb, sun2019deep}.
Domain adaptive keypoint detection \cite{jiang2021regressive, ohkawa2022domain, zhang2019bridging, Cao_2019_ICCV, kundu2022uncertainty, doersch2019sim2real, lee2022uda} attempts to address this issue by transferring information from a labeled source domain to an unlabeled target domain, avoiding the extra annotation costs.
These methods are primarily divided into 3D and 2D keypoint detection.
In 3D keypoint detection, this problem is complex and there are many prior works.
For instance, \cite{doersch2019sim2real} proves that the neural networks can perform well when the data is pre-processed to extract cues about the person’s motion, notably as optical flow and the motion of 2D keypoints, and therefore propose to use motion as a simple way to bridge a Synthetic-to-realistic gap when the video is available.
UDA-COPE \cite{lee2022uda} introduces a bidirectional filtering method between the predicted normalized object coordinate space (NOCS) map and observed point cloud, to promote teacher-student consistency training.

This paper focuses on another type, \ie, 2D keypoint detection.
In this field, existing methods can be primarily divided into human keypoint detection and animal keypoint detection.
For human keypoint detection, RegDA \cite{jiang2021regressive}, and MarsDA \cite{jin2022multi} introduce adversarial regressors to narrow the domain gap.
C-GAC \cite{ohkawa2022domain} integrates the proposed prediction confidence into self-training to obtain reliable pseudo labels. 
For animal keypoint detection, CC-SSL \cite{mu2020learning}, and UDA-Animal \cite{li2021synthetic} are based on transformation consistency and the pseudo-label refinery technique.
Besides, unified domain adaptive pose estimation (UDAPE) \cite{kim2022unified} proposes to align representations using both input-level and output-level cues, and provides a unified framework for both human keypoint detection and animal keypoint detection problems.
Generally, the source data is essential for these carefully designed alignment strategies, which may violate data privacy law and pose serious concerns, while our method solves this problem under a source data-free setting.

\subsection{Curriculum Learning}
Curriculum learning (CL) \cite{bengio2009curriculum} is a training strategy that trains a model from easy to hard, which simulates the learning principle of humans and animals.
CL serves as a general training strategy and has been used in multiple applications of CL in machine learning \cite{fan2018learning, graves2017automated, hacohen2019power, wang2021survey}.
In these applications, the motivations can be categorized for applying CL into two groups: to guide, regularizing the training towards better regions in parameter space (with steeper gradients) from the perspective of the optimization problem, and to denoise, focusing on the high-confidence easier area to alleviate the interference of noisy data as from the perspective of data distribution.
A general CL framework has a Difficulty Measurer, which decides the relative “easiness” of each data example, and a Training Scheduler, which decides the sequence of data subsets throughout the training process based on the judgment from the Difficulty Measurer.
Based on this framework, CL is primarily divided into two cases, \ie, predefined CL and automatic CL.
The predefined CL case measures the difficulty of samples and schedules the training process based on prior knowledge, and the automatic CL case adopts a data-driven method to learn any (or both) aspects.
In predefined CL, researchers manually design various Difficulty Measurers mainly based on the data characteristics of specific tasks. For instance, in NLP tasks, sentence length is the most popular Difficulty Measure in NLP tasks \cite{platanios2019competence}. 
This is because researchers intuitively think the complexity of a sentence or paragraph can be expressed by sentence length.

Self-paced learning (SPL) \cite{kumar2010self, tullis2011effectiveness} is a typical example of automatic curriculum learning.
The easiest samples with the lowest training loss are used to train the models, this portion of the easiest samples gradually increases according to a specified scheduler.
Based on the scheduler, the models are able to learn at their own pace, including deciding what, how, when, and how long to study.
Prior work formulates the critical principle of SPL as a concise model \cite{kumar2010self}, and proves that the SPL regime is convergent by adopting an alternative optimization strategy (AOS) \cite{meng2015objective}.
Generally, previous self-paced learning methods focus on the supervised learning scenario, and no few works study the application of SPL on unsupervised learning tasks. 
Traditional supervised SPL achieves promising results, while unsupervised SPL still remains much room for improvement. 
In this paper, we study unsupervised SPL in regression problems.

\begin{figure*}[t]
\centering
\includegraphics[width=2\columnwidth]{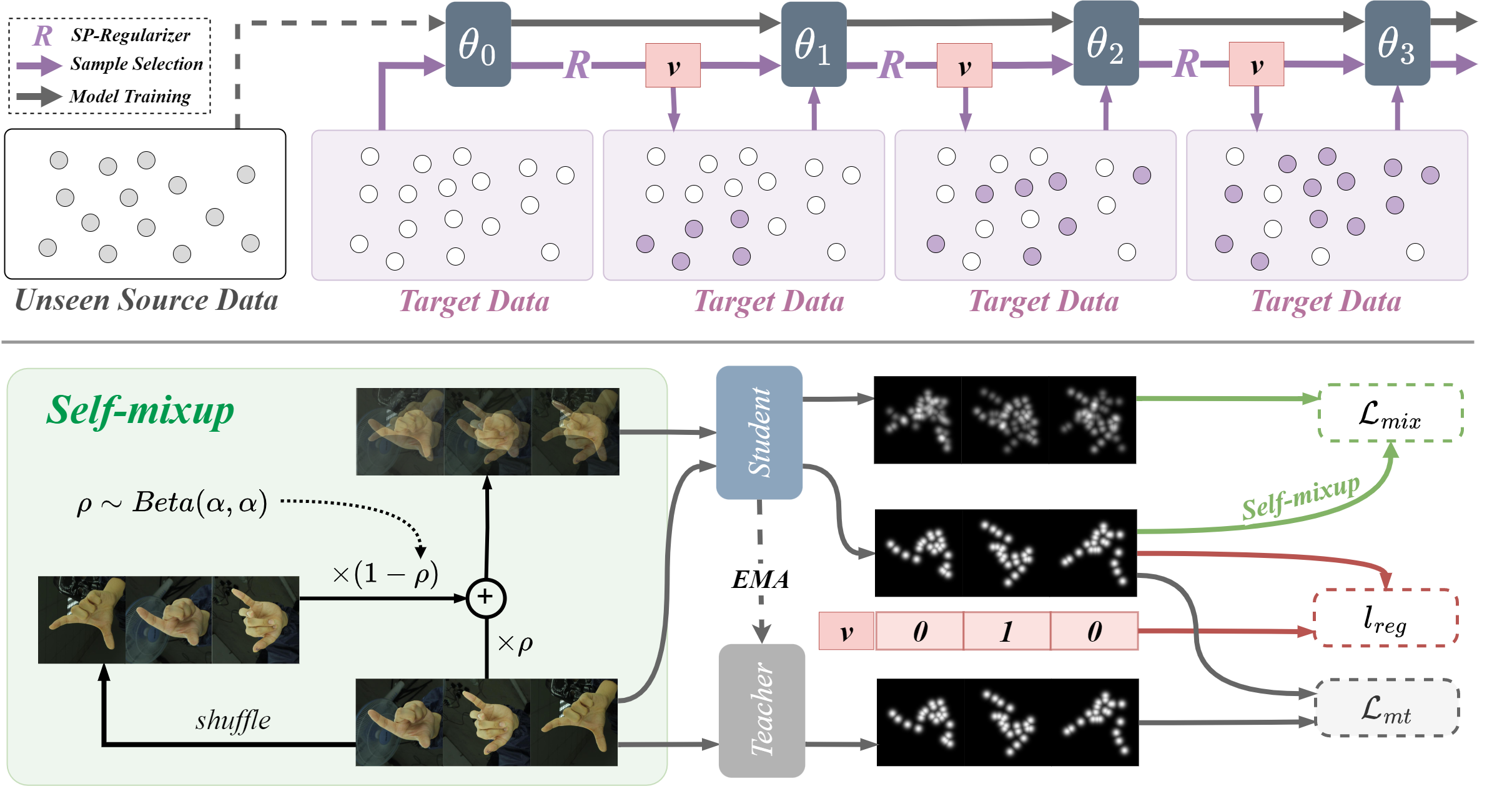}
\caption{
    The pipeline of Mixup Augmentation and Progressively Selection (MAPS). 
    Given the well-trained source model $\theta_0$, MAPS progressively selects easy samples from target samples, \ie, assigns weight, 1 for easy samples and 0 for the remaining samples, the weights are notated as symbol $v$. 
    Then these samples participate in the model updating with self-mixup training, which regularizes the model to favor a linear behavior in-between the target samples. 
    These two aspects are alternatively performed until the loss function converges.
    }
\label{pipeline}
\end{figure*}

\section{Method}
We define the notations in Section \ref{sec_not} and generate the source model by a standard regression loss in section \ref{sec_src}.
Then we construct a simple baseline by the mean-teacher framework in Section \ref{sec_mt} and introduce the proposed noise-robustness progressive learning method in Section \ref{sec_maps}.

\subsection{Notations}\label{sec_not}
For a standard unsupervised domain adaptive 2D keypoint detection, we have $n_s$ labeled samples $\{x^s_i,y^s_i\}_{i=1}^{n_s}$ from the source domain $D_s = \{\mathcal{X}_s, \mathcal{Y}_s^K\}$, and $n_t$ unlabeled data $\{x^t_i\}_{i=1}^{n_t}$ from the target domain $D_t = \{\mathcal{X}_t\}$, 
where $\mathcal{X}_s,\mathcal{X}_t \in \mathcal{R}^{H\times W\times 3}$ are the input space, $\mathcal{Y}_s^K \in \mathcal{R}^2$ is the output space, $K$ is the number of keypoints in each input image.
The goal is to find a keypoint detector to predict the labels $\{y^t_i\}_{i=1}^{n_t}$ of the target domain.

The source-free unsupervised domain adaptive keypoint detection task aims to learn a target keypoint detector $f_t:\mathcal{X}_t\to\mathcal{Y}_t$ and infer  $\{y^t_i\}_{i=1}^{n_t}$ with only $\{x^t_i\}_{i=1}^{n_t}$ and the source keypoint detector $f_s$.
We first generate the source keypoint detection model by a standard regression method, and then we transfer the model to the target domain without accessing the source data.

\subsection{Source Model Generation}\label{sec_src}
We consider to train the source keypoint detection model $f_s:\mathcal{X}_s\to \mathcal{Y}_s$ by minimizing the following standard supervised regression loss,
\begin{equation}
   \mathcal{L}_{src}(\mathcal{X}_s, \mathcal{Y}_s) = \sum\limits_{i=1}^{n_s}\|f_s(x_i^s)-H(y_i^s)\|_2,
\end{equation}
where $H(\cdot)$ denotes the heatmap generation function \cite{tompson2014joint}.
To further improve the performance of the source model and facilitate the following target data alignment, we use the random data augmentation technique \cite{li2021synthetic} at this stage.
Taking the augmentation into consideration, the objective function is reformulated as:
\begin{equation}
\begin{aligned}
    \mathcal{L}^{aug}_{src}(\mathcal{X}_s, \mathcal{Y}_s) =\sum\limits_{i=1}^{n_s}\|f_s(\mathcal{A}^s(x_i^s))-\mathcal{A}^s(H(y_i^s))\|_2,
\end{aligned}
\end{equation}
where $\mathcal{A}^s$ denotes the augmentation function.

\subsection{Teacher-Student Consistency Learning}\label{sec_mt}
Given the well-trained source model $f_s$, we first construct a baseline method based on the mean-teacher framework \cite{tarvainen2017mean, li2021synthetic, kim2022unified}.
Specifically, the student model $f_t$ and teacher model $f'_t$ have an identical architecture with the given source model $f_s$, and the teacher parameter $\theta '$ is updated with the exponential moving average (EMA) of the student parameter $\theta$:
\begin{equation}
    \theta'_t=\eta \theta'_{t-1} + (1-\eta) \theta_t,\label{ema}
\end{equation}
where $t$ denotes the step of training, and $\eta$ denotes the smoothing coefficient which is set to 0.999 by default.

Here we also introduce the random data augmentation function $A^t$ on target data, the input of student $\mathcal{A}^t_1(x_i^t) $ and the input of teacher $\mathcal{A}^t_2(x_i^t) $ are generated from different augmentation $\mathcal{A}^t_1$ and $\mathcal{A}^t_2$, sampled from $\mathcal{A}^t$.
Besides, considering the missing and occluded keypoints in some images, we only force the teacher-student consistency on the predicted points with the maximum activation greater than a threshold $\tau$.
The object function is formulated as:
\begin{equation}
\begin{aligned}
    \mathcal{L}_{mt}(\mathcal{X}_t)=\sum\limits_{i=1}^{n_t} \sum\limits_{k=1}^{K}\mathbbm{1}(h'_k\geq \tau)\|\Tilde{\mathcal{A}^t_2}(h'_k)-\Tilde{\mathcal{A}^t_1}(h_k)) \|_2,\label{lmt}
\end{aligned}
\end{equation}
where $\Tilde{A}^t_1$ and $\Tilde{A}^t_2$ denote the inverse function of $\mathcal{A}^t_1$ and $\mathcal{A}^t_2$, 
$\{h_k\}_{k=1}^K=f_t(\mathcal{A}^t_1(x_i^t))$ denote the outputs of student model, 
$\{h'_k\}_{k=1}^K=H(f'_t(\mathcal{A}^t_2(x_i^t))$ denote the normalized outputs of teacher model, 
$\mathbbm{1}(\cdot)$ denotes the indicator function, and $K$ denotes the number of keypoints.

\subsection{Noise-Robust Progressive Learning}\label{sec_maps}
Based on the baseline depicted in section \ref{sec_mt}, we further employ a simple strategy, pseudo-labeling \cite{lee2013pseudo}, which directly uses the current prediction of the model as the ground truth, to explore the semantic information in the target domain.
However, this manner inevitably introduces many erroneous labels, making the model suffer from noise accumulation, which obstructs true signal and biasing estimation of corresponding parameters \cite{elman2020noise}, leading to sub-optimal results.
As shown in Fig. \ref{pipeline}, we further propose Mixup Augmentation and Progressive Selection (MAPS), to avoid the over-confidence in outliers and select reliable pseudo labels progressively to participate in the model training.

\noindent
\textbf{Mixup Augmentation.}
Based on the baseline model, we propose self-mixup augmentation to favor simple linear behavior in-between target data during training, thereby preventing confirmation bias \cite{zhang2017mixup} and improving the model's robustness.
To be specific, we mix the augmented target samples $\mathcal{A}^t(x_i^t)$ and another random sample $\mathcal{A}^t(x_j^t)$ ($i\neq j$) in the current mini-batch,  with a mix ratio $\rho\sim Beta(\alpha, \alpha)$ to construct the self-mixup augmentation, and then we introduce the self-mixup loss in the following:
\begin{equation}
\begin{aligned}
    \mathcal{L}_{mix} (\mathcal{X}_t) &=\sum\limits_{i=1}^{n_t}\|f_t(x_i^m) - h_i^m\|_2, \\
    x_i^m&=\rho \mathcal{A}^t(x_i^t) + (1-\rho)\mathcal{A}^t(x_j^t), \\
     h_i^m&=\rho f_t(\mathcal{A}^t(x_i^t))+(1-\rho) f_t(\mathcal{A}^t(x_j^t)).\label{lmix}
\end{aligned}
\end{equation}
The self-mixup loss regularizes the model to favor simple linear behavior in-between the target samples, preventing the over-fitting of outliers.

\noindent
\textbf{Progressive Selection.}
The pseudo-labeling technique \cite{lee2013pseudo} picks up the current prediction as used as if they were true labels. 
However, it is inevitable to introduce noisy labels, which might mislead the model over time.
We introduce the self-paced learning regime \cite{kumar2010self} to progressively select samples with high-confident pseudo labels, and the pseudo-label regression loss is imposed on them.
This aim can be realized by minimizing the following objective:
\begin{equation}
\begin{aligned}
    \mathcal{L}_{spl}(\mathcal{X}_t) = \sum\limits_{i=1}^{n_t} v_i l_{reg}(x_i^t) + R(v_i,\lambda),\ s.t.\ v_i\in[0,1]\\
\end{aligned}
\end{equation}
where $\lambda$ denotes the age parameter for adjusting the learning pace, 
$v_i$ denotes the weight for the $i$-th sample, 
$R(v_i,\lambda)$ denotes the self-paced regularizer (SP-regularizer), 
and $l_{spl}$ denotes the self-paced loss weighted by $v_i$. 
The self-paced loss for each sample $x_i^t$ is defined by the mean squared error, \ie, pseudo-label regression loss:
\begin{equation}
l_{reg} (x_i^t) = \|f_t(\mathcal{A}^t(x_i^t))-\mathcal{A}^t(H(\hat{y}_i^t))\|_2,
\label{lspl1}
\end{equation}
where $\hat{y}_i^t$ is the pseudo label of $x_i^t$, which is the 2D coordinate of the maximum activation of the student output $f_t(x_i^t)$.

Here we adopt the hard SP-regularizer \cite{kumar2010self} to solve the sample weights. 
The formulation and the closed-form solution are below:
\begin{equation}
    R(v_i,\lambda)=-\lambda v_i;\  v_i=\left\{
    \begin{aligned}
    1, \ l_{reg}(x_i^t)<\lambda \\
    0, \ l_{reg}(x_i^t)\geq \lambda\label{sp-reg}
    \end{aligned}
    \right. .
\end{equation}
The model parameters are fixed during selection, therefore, $l_{reg}$ can be seen as a difficulty measurer, and the age parameter $\lambda$ is a threshold.
This regime progressively selects the easy samples by assigning the weight 1 for them.

The age parameter $\lambda$ is updated by a baby step scheduler $S$ \cite{li2017self}.
To be specific, we pre-defines an increasing sequence $M=\{M_1, M_2, M_3,..., M_Q\}$.
In the $q$-th round, we select $M_q$ samples, and $\lambda$ can be calculated by:
\begin{equation}
\lambda= S\big({M_{q}},\{l_{reg}(x_i^t)\}_{i=1}^{n_t}\big), \label{babystep}
\end{equation}
where $S(m,L)$ denotes the $m$-th smallest value of loss function $L$.

\noindent
\textbf{Overall Objective.}
With the above two aspects, the overall objective of MAPS can be summarized as:
\begin{equation}
\begin{aligned}
     \mathop{\min}_{\theta,v}\ \mathcal{L}_{mt}(\mathcal{X}_t)+\beta_{m}\mathcal{L}_{mix}(\mathcal{X}_t)+\beta_{s} \mathcal{L}_{spl}(\mathcal{X}_t), \label{loverall}
\end{aligned}
\end{equation}
where $\beta_{s}$,  $\beta_{m}$ are hyper-parameters to control the weights of the corresponding loss terms.
We adopt the alternative optimization strategy to optimize the objective function in the following steps.

1) Select easy samples, \ie, solve weights $v$, when $\theta$ and $\lambda$ are fixed. Optimizing Eq. (\ref{loverall}) is equal to solving the following problem:
\begin{equation}
\begin{aligned}
    \mathop{\min}_{v} \ \mathcal{L}_{spl}(\mathcal{X}_t),\  s.t. \ v_i \in [0,1],\label{s1}
\end{aligned}
\end{equation}
we adopt the hard SP-regularizer here, and the closed-form solution of $v$ is shown in Eq. (\ref{sp-reg}).

2) Train the model, \ie, update $\theta$, when $v$ and $\lambda$ are fixed.
Optimizing Eq. (\ref{loverall}) is equal to solving the following problem:
\begin{equation}
\begin{aligned}
    \mathop{\min}_{\theta} \mathcal{L}_{mt}(\mathcal{X}_t)+\beta_{m}\mathcal{L}_{mix}(\mathcal{X}_t)+\beta_{s}\sum\limits_{i=1}^{n_t}v_i l_{reg}(x_i^t).\label{s2}
\end{aligned}
\end{equation}

In addition, according to the given baby step scheduler in Eq. (\ref{babystep}), we also update the age parameter $\lambda$ to include more training samples during the optimization process. The overall algorithm is shown in Algorithm \ref{alg1}.
\begin{algorithm}
	\renewcommand{\algorithmicrequire}{\textbf{Input:}}
	\renewcommand{\algorithmicensure}{\textbf{Output:}}
	\caption{Algorithm of the proposed MAPS.}
	\label{alg1}
	\begin{algorithmic}[1]
	    \REQUIRE Target dataset $\mathcal{X}_t$; a well-trained source model $f_s$;
	    \ENSURE New target model $f_t$ and $f'_t$;
	    \STATE Initialize the student $f_t$ and teacher $f'_t$ with $f_s$;
		\REPEAT
		\STATE Fix $f_t$, update $v$ by Eq. (\ref{s1}); // sample selection
		\STATE Update $f_t$ by Eq. (\ref{s2}); // model updating
		\STATE Update $f'_t$ by Eq. (\ref{ema});
		\STATE Update $\lambda$ based on $S$ by Eq. (\ref{babystep});
		\UNTIL Progressive selection rounds are exhausted.
	\end{algorithmic}
\label{algo}
\end{algorithm}

\section{Experiment}
We first give the experimental setup in Section \ref{sec-setup}, then we conduct the experiments and analyze the results on hand keypoint detection, human keypoint detection, and animal keypoint detection in Section \ref{sec-hand}, \ref{sec-human}, \ref{sec-animal}, respectively.
To validate the effectiveness of the proposed modules, we conduct the ablation study in Section \ref{sec-ablation} and analyze the sensitivity in Section \ref{sec-further}.

\subsection{Setup}\label{sec-setup}
The architecture of our model is based on Simple Baseline \cite{xiao2018simple} with the backbone of pre-trained ResNet101 \cite{he2016deep}. 
Following prior works \cite{kim2022unified, li2021synthetic, mu2020learning}, we add texture and geometric augmentation including Gaussian noise, Gaussian blurring, rotation, and random 2D translation for both source model generation and target model training.
The parameter $\alpha$ of the beta distribution is set to 0.75, and the weights of loss functions $\beta_{m}$ and $\beta_{s}$ are both empirically set to 1.
In the source model generation stage, we adopt Adam \cite{kingma2014adam} as the optimizer and set the learning rate as 2e-4, and it decreased to 2e-5 at 15,000 steps and 2e-6 at 20,000 steps, while there is a total of 25,000 steps in this stage.
In the target model training, we also adopt the Adam optimizer and learning rate 2e-4, and the learning rate decreases to 1e-4 at 2,500 steps with a total of 7,500 steps.
Our basic augmentation is based on UDAPE \cite{kim2022unified}, which adds rotation and random 2D translation on the augmentation in RegDA \cite{jiang2021regressive}.
We use the Percentage of Correct Keypoints (PCK) as our metric, in which estimation is considered correct if its distance from the ground truth is less than a fraction of 0.05 of the image size.
We randomly run our methods three times with different random seeds \{0,1,2\} via \textbf{PyTorch} and report the average accuracies.

We conduct experiments on three tasks with four different target datasets, and compare MAPS with several state-of-the-art \emph{non-source-free} methods, including the animal keypoint detection methods CCSSL \cite{mu2020learning} and UDA-Animal \cite{li2021synthetic}; the human and hand keypoint detection method RegDA \cite{jiang2021regressive}, and the unified method UDAPE \cite{kim2022unified}.
In the table of comparative experiments, `source only' denotes using the source model for prediction, `MT' is our baseline method introduced in section \ref{sec_mt}, and MAPS is the full version of our method.
All the results except ours are from the original paper of UDAPE, and the dataset split policy follows UDAPE and RegDA.

\begin{table}
    \centering 	
    \caption{PCK@0.05 on task \textbf{RHD}$\to$\textbf{H3D}. (Best value is in \textcolor{RedOrange}{\textbf{Redorange}} color. Second best value is in \textcolor{Periwinkle}{\textbf{periwinkle}} color.)}
    \resizebox{0.46\textwidth}{!}{     	 
    \begin{tabular}{lccccca}
    \toprule
    Method & SF & MCP & PIP & DIP & Fin & Avg. \\
    \midrule
    CCSSL \cite{mu2020learning} &$\times$ & 81.5	&79.9	&74.4	&64.0	&75.1 \\ 
    UDA-Animal \cite{li2021synthetic}&$\times$ &82.3&	79.6&	72.3&	61.5&	74.1 \\
    RegDA \cite{jiang2021regressive} &$\times$& 79.6&	74.4&	71.2&	62.9&	72.5 \\
    UDAPE\cite{kim2022unified} &$\times$&\secondbest{86.7}&	\textcolor{Periwinkle}{\textbf{84.6}}&	\textcolor{Periwinkle}{\textbf{78.9}}&	\textcolor{Periwinkle}{\textbf{68.1}}&	\textcolor{Periwinkle}{\textbf{79.6}} \\
    \midrule
    Source only & -&  65.0&	62.4&	61.1&	54.0&	60.1\\
    MT & $\checkmark$ & 85.0	&\textcolor{Periwinkle}{\textbf{84.6}}&	77.8&	67.3&	78.8\\
    MAPS & $\checkmark$ & \best{86.9} & \best{84.8} &	\best{79.1} &	\best{69.3} &	\best{80.0} \\
    \bottomrule 
    \end{tabular} 
    }
    \label{h3d} 
\end{table} 

\begin{figure*}[t]
\centering
\includegraphics[width=1.8\columnwidth]{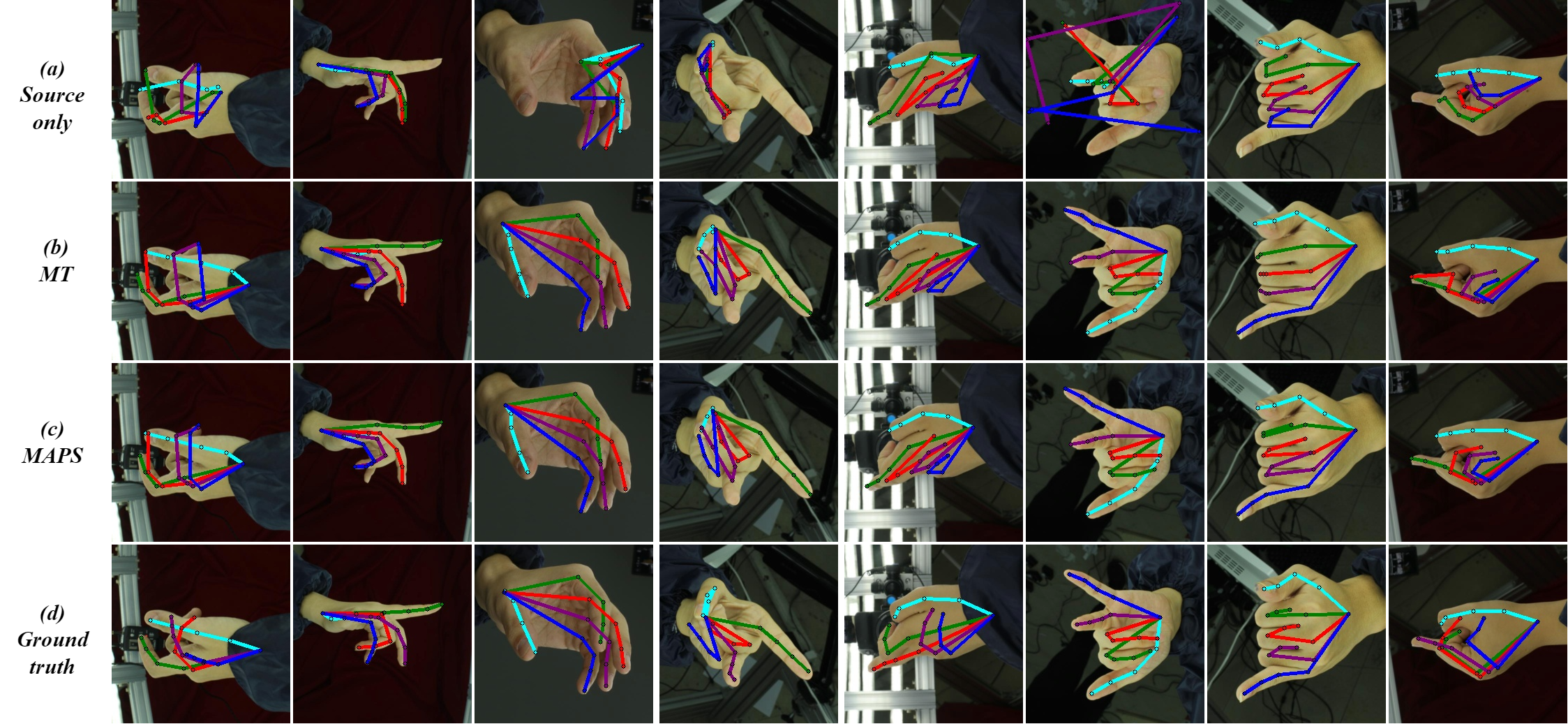}
\caption{Qualitative results of task \textbf{RHD}$\to$\textbf{H3D}.}
\label{vis-hand}
\end{figure*}

\begin{table*}
\centering
\caption{PCK@0.05 on task \textbf{SURREAL}$\to$\textbf{LSP}.}
\resizebox{0.65\textwidth}{!}{
\begin{tabular}{lccccccca}
    \toprule
    Method & SF & Sld& Elb& Wrist& Hip& Knee& Ankle & Avg. \\
    \midrule
    CCSSL \cite{mu2020learning} & $\times$ & 36.8&	66.3&	63.9&	59.6&	67.3&	70.4&	60.7 \\ 
    UDA-Animal \cite{li2021synthetic}&$\times$ & 61.4&	77.7&	75.5&	65.8&	76.7&	78.3&	69.2\\
    RegDA \cite{jiang2021regressive} &$\times$ & 62.7&	76.7&	71.1&	81.0&	80.3&	75.3&	74.6 \\
    UDAPE\cite{kim2022unified} &$\times$ & \textcolor{RedOrange}{\textbf{69.2}}&	\textcolor{RedOrange}{\textbf{84.9}}&	\textcolor{RedOrange}{\textbf{83.3}}&	\textcolor{RedOrange}{\textbf{85.5}}&	\textcolor{RedOrange}{\textbf{84.7}}&	\textcolor{Periwinkle}{\textbf{84.3}}&	\textcolor{RedOrange}{\textbf{82.0}}\\
    \midrule
    Source only &-&  50.6&	64.8&	63.3&	70.1&	71.2&	70.1&	65.0\\
    MT &$\checkmark$ & \textcolor{RedOrange}{\textbf{69.2}}&	82.1&	80.2&	83.4&	82.3&	80.8&	79.6\\
    MAPS &$\checkmark$&67.0	&\textcolor{Periwinkle}{\textbf{84.2}}&	\textcolor{Periwinkle}{\textbf{82.7}}&	\textcolor{Periwinkle}{\textbf{84.0}}	&\textcolor{Periwinkle}{\textbf{83.8}}&	\textcolor{RedOrange}{\textbf{84.6}}&	\textcolor{Periwinkle}{\textbf{81.0}}\\
    \bottomrule 
\end{tabular}
}
\label{lsp}
\end{table*}
\begin{table*}
\centering
\caption{PCK@0.05 on task \textbf{SynAnimal}$\to$\textbf{AnimalPose}.}
    \resizebox{0.8\textwidth}{!}{
    \begin{tabular}{lc|cccca|cccca}     
    \toprule     
    \multirow{2}{*}{Method} & \multirow{2}{*}{SF} &\multicolumn{5}{c|}{Dog} &\multicolumn{5}{c}{Sheep} \\      
    \cmidrule{3-12}      &&Eye&	Hoof&	Knee&	Elb&	Avg.&	Eye&	Hoof&	Knee&	Elb&	Avg. \\     
    \midrule     
    CCSSL \cite{mu2020learning} &$\times$ & 34.7&	37.4&	25.4&	19.6&	27.0&	44.3&	55.4&	43.5&	28.5&	42.8 \\      
    UDA-Animal \cite{li2021synthetic}&$\times$ &26.2&	39.8&	31.6&	24.7&	31.1&	48.2&	52.9&	49.9&	29.7&	44.9 \\     
    RegDA \cite{jiang2021regressive} &$\times$ & 46.8&	54.6&	32.9&	31.2&	40.6&	62.8&	68.5&	57.0&	42.4&	56.9 \\     
    UDAPE\cite{kim2022unified} &$\times$ &\textcolor{Periwinkle}{\textbf{56.1}}&	59.2&	\textcolor{RedOrange}{\textbf{38.9}}&	\textcolor{Periwinkle}{\textbf{32.7}}&	\textcolor{Periwinkle}{\textbf{45.4}}&	61.6&	\textcolor{Periwinkle}{\textbf{77.4}}&	57.7&	\textcolor{RedOrange}{\textbf{44.6}}&	60.2 \\     
    \midrule     
    Source only &-& 38.2&	43.2&	25.7&	24.1&	32.0&	59.9&	60.7&	46.2&	31.0&	47.9 \\     
    MT & $\checkmark$ & 53.0	&\textcolor{Periwinkle}{\textbf{59.3}}&	\textcolor{Periwinkle}{\textbf{37.7}}&	32.0&	44.4&	\textcolor{Periwinkle}{\textbf{66.1}}&	76.5&	\textcolor{RedOrange}{\textbf{59.8}}&	42.6&	\textcolor{Periwinkle}{\textbf{60.6}} \\     
    MAPS &$\checkmark$ & \textcolor{RedOrange}{\textbf{63.2}}&	\textcolor{RedOrange}{\textbf{60.5}}&	37.3&	\textcolor{RedOrange}{\textbf{34.0}}&	\textcolor{RedOrange}{\textbf{46.7}}&	\textcolor{RedOrange}{\textbf{74.2}}&	\textcolor{RedOrange}{\textbf{77.9}}&	\textcolor{Periwinkle}{\textbf{58.8}}&	\textcolor{Periwinkle}{\textbf{43.3}}&	\textcolor{RedOrange}{\textbf{62.0}}\\     
    \bottomrule
    \end{tabular}  
    }
\label{animalpose}
\end{table*}
\subsection{Hand Keypoint Detection}\label{sec-hand}
\noindent
\textbf{Dataset.} 
Rendered Hand Pose Dataset \cite{zimmermann2017learning} (\textbf{RHD}) is a synthetic dataset containing 41,258 training images and 2,728 testing images with corresponding 21 hand keypoints labels. 
Hand-3D-Studio \cite{zhao2020hand} (\textbf{H3D}) is a real-world multi-view hand image dataset containing 22k images in total. 
Following the split way in RegDA \cite{jiang2021regressive}, we select 3.2k images as the testing set, and the remaining as the training set.

\noindent
\textbf{Implementation Details.}
We use \textbf{RHD} as the source domain and \textbf{H3D} as the target domain.
The number of progressive selection rounds $Q$ and the increasing sequence $M$ in the baby step schedular are set to $3$ and $\{0.25N,0.35N,0.45N\}$, respectively, where $N$ is the total number of training samples.
We report 21 keypoints on the different anatomical parts of a hand including metacarpophalangeal (MCP), proximal interphalangeal (PIP), distal interphalangeal (DIP), and fingertip (Fin).

\noindent
\textbf{Results.}
The quantitative results are presented in Table \ref{h3d}.
In comparison, MAPS achieves the highest average accuracy even compared with the non-source-free methods, and MT also achieves competitive results.
We outperform a large margin than most of the methods except UDAPE \cite{kim2022unified}, demonstrating the effectiveness of our method.
MAPS also outperforms MT, as the noise accumulation phenomenon in MT is reduced.
We show the qualitative results in Fig. \ref{vis-hand}.
The results of `source-only' are not satisfied in some complicated hand poses due to a large domain gap, and MT successfully aligns the target domain to the unseen source domain. 
MAPS further refines the results of MT, obtaining more accurate results.

\begin{figure*}[t]
\centering
\includegraphics[width=1.8\columnwidth]{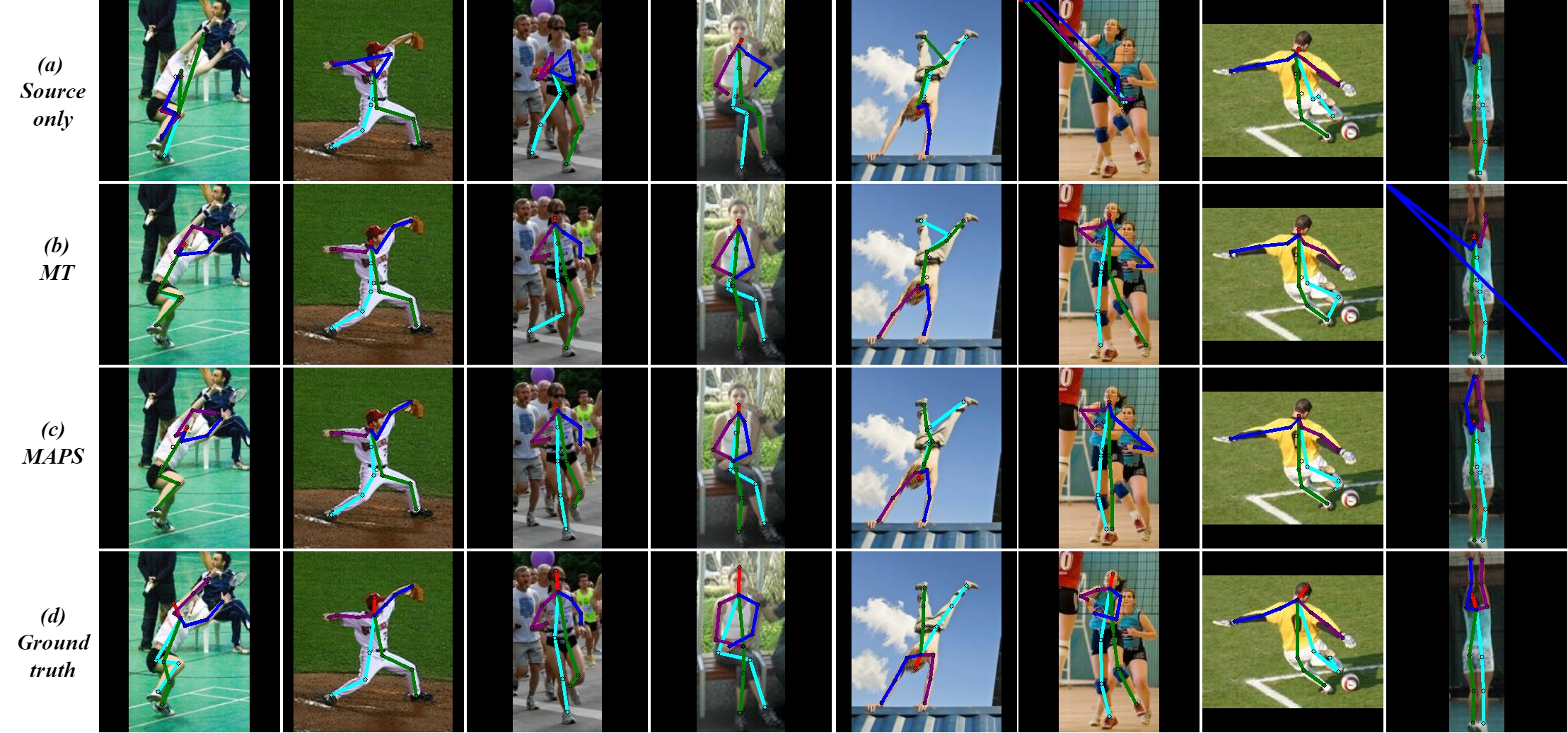}
\caption{Qualitative results of task \textbf{SURREAL}$\to$ \textbf{LSP}.}
\label{vis-lsp}
\end{figure*}
\begin{table*}
    \centering 	
    \caption{PCK@0.05 on task \textbf{SynAnimal}$\to$\textbf{TigDog}.}
    \resizebox{1\textwidth}{!}{
    \begin{tabular}{lc|ccccccca|ccccccca}
    \toprule
    \multirow{2}{*}{Method} & \multirow{2}{*}{SF} &\multicolumn{8}{c|}{Horse} &\multicolumn{8}{c}{Tiger} \\ 
    \cmidrule{3-18}
    & & Eye&	Chin&	Sld&	Hip&	Elb&	Knee&	Hoof&	Avg.&	Eye&	Chin&	Sld&	Hip&	Elb&	Knee&	Hoof&	Avg. \\
    \midrule
    CCSSL \cite{mu2020learning} &$\times$ & \textcolor{Periwinkle}{\textbf{89.3}}&	\textcolor{Periwinkle}{\textbf{92.6}}&	69.5&	78.1&	70&	73.1&	65&	73.1&	94.3&	91.3&	49.5&	70.2&	\textcolor{RedOrange}{\textbf{53.9}}&	59.1&	70.2&	66.7 \\ 
    UDA-Animal \cite{li2021synthetic}&$\times$ &86.9&	\textcolor{RedOrange}{\textbf{93.7}}&	\textcolor{RedOrange}{\textbf{76.4}}&	\textcolor{RedOrange}{\textbf{81.9}}&	70.6&	\textcolor{RedOrange}{\textbf{79.1}}&	\textcolor{RedOrange}{\textbf{72.6}}&	77.5&	98.4&	87.2&	49.4&	\textcolor{RedOrange}{\textbf{74.9}}&	49.8&	\textcolor{Periwinkle}{\textbf{62}}&	\textcolor{RedOrange}{\textbf{73.4}}&	\textcolor{Periwinkle}{\textbf{67.7}} \\
    RegDA \cite{jiang2021regressive} &$\times$ & 89.2&	92.3&	70.5&	77.5&	71.5&	72.7&	63.2&	73.2&	93.3&	92.8&	50.3&	67.8&	50.2&	55.4&	60.7&	61.8 \\
    UDAPE\cite{kim2022unified} &$\times$ &\textcolor{RedOrange}{\textbf{91.3}}&	92.5&	74.0&	74.2&	\textcolor{RedOrange}{\textbf{75.8}}&	\textcolor{Periwinkle}{\textbf{77.0}}&	\textcolor{Periwinkle}{\textbf{66.6}}&	\textcolor{RedOrange}{\textbf{76.4}}&	98.5&	\textcolor{RedOrange}{\textbf{96.9}}&	\textcolor{RedOrange}{\textbf{56.2}}&	63.7&	\textcolor{Periwinkle}{\textbf{52.3}}&	\textcolor{RedOrange}{\textbf{62.8}}&	\textcolor{Periwinkle}{\textcolor{RedOrange}{\textbf{72.8}}}&	\textcolor{RedOrange}{\textbf{67.9}} \\
    \midrule
    Source only &- &  82.0 &	90.0 &	59.2 &	\textcolor{Periwinkle}{\textbf{79.5}} &	65.8 &	66.9 &	57.7 &	67.4 &	85.4 &	81.8 &	44.6 	&\textcolor{Periwinkle}{\textbf{70.8}} &	39.6 &	48.4 &	55.5 &	54.8  \\
    MT &$\checkmark$ & 80.9 &	90.7 &	71.3 &	75.5 &	\textcolor{Periwinkle}{\textbf{72.9}} &	74.0 &	66.1 &	73.5 &	\textcolor{Periwinkle}{\textbf{98.7}} &	91.5 &	47.9 &	56.5 &	48.2 &	60.5 &	68.3 &	63.9 \\
    MAPS &$\checkmark$ & 82.3 &	91.4 &	\textcolor{Periwinkle}{\textbf{74.1 }}&	76.5 &	71.8 &	74.2 &	65.3 &	\textcolor{Periwinkle}{\textbf{73.7}} &	\textcolor{RedOrange}{\textbf{99.5}} &	\textcolor{Periwinkle}{\textbf{93.8}} &	\textcolor{Periwinkle}{\textbf{55.1}} &	60.7 &	48.7 &	60.9 &	71.4 &	66.0  \\
    \bottomrule 
    \end{tabular} 
    }
    \label{tigdog} 
\end{table*}

\begin{table*}
    \centering 	
    \caption{The effectiveness of three loss functions and their interactions on task \textbf{SynAnimal}$\to$\textbf{AnimalPose}. }
    \resizebox{0.8\textwidth}{!}{     	 
    \begin{tabular}{ccc|cccca|cccca}
    \toprule
    \multirow{2}{*}{$\mathcal{L}_{mt}$} &\multirow{2}{*}{$\mathcal{L}_{mix}$}&\multirow{2}{*}{$\mathcal{L}_{spl}$}&\multicolumn{5}{c|}{\textbf{SynAnimal}$\to$\textbf{AnimalPose}: Dog} &\multicolumn{5}{c}{\textbf{SynAnimal}$\to$\textbf{AnimalPose}: Sheep}\\
    \cmidrule{4-13}
    &&& Eye&	Hoof&	Knee&	Elb	&Avg. & Eye&	Hoof&	Knee&	Elb&	Avg.\\
    \midrule
    && &38.2	&43.2&	25.7&	24.1&	32.0&	59.9&	60.7&	46.2&	31.0&	47.9\\
    $\checkmark$ &&&53.0&	59.3&	37.7&	32.0	&44.4&	66.1&	76.5&	\textcolor{RedOrange}{\textbf{59.8}}&	42.6&	60.6\\
    $\checkmark$ & $\checkmark$ &&56.6&	59.5&	\textcolor{RedOrange}{\textbf{39.2}}&	33.2&	45.8&	69.5&	\textcolor{RedOrange}{\textbf{78.2}}&	59.3&	42.6&	61.4\\
    $\checkmark$ & & $\checkmark$ &59.7	&59.1&	36.7&	32.5&	45.2&	72.3&	76.8&	58.9&	\textcolor{RedOrange}{\textbf{43.6}}&	61.6\\
    $\checkmark$ & $\checkmark$ & $\checkmark$ &\textcolor{RedOrange}{\textbf{63.2}}&	\textcolor{RedOrange}{\textbf{60.5}}&	37.3&	\textcolor{RedOrange}{\textbf{34.0}}&	\textcolor{RedOrange}{\textbf{46.7}}&	\textcolor{RedOrange}{\textbf{74.2}}&	77.9&	58.8&	43.3&	\textcolor{RedOrange} {\textbf{62.0}}\\
    \bottomrule 
    \end{tabular} 
    }
    \label{ablation-animal} 
\end{table*} 
\begin{table*}
    \centering 	
    \caption{The effectiveness of three loss functions and their interactions on task \textbf{SURREAL}$\to$\textbf{LSP}. }
    \resizebox{0.7\textwidth}{!}{     	 
    \begin{tabular}{ccc|cccccca}
    \toprule
    \multirow{2}{*}{$\mathcal{L}_{mt}$} &\multirow{2}{*}{$\mathcal{L}_{mix}$}&\multirow{2}{*}{$\mathcal{L}_{spl}$}& \multicolumn{7}{c}{\textbf{SURREAL}$\to$\textbf{LSP}}\\
    \cmidrule{4-10}
    &&& Sld& Elb &Wrist& Hip& Knee& Ankle& Avg. \\
    \midrule
    && & 50.6&	64.8&	63.3&	70.1&	71.2&	70.01	&65.0\\
    $\checkmark$ &&&	\textcolor{RedOrange}{\textbf{69.2}}&	82.1&	80.2&	83.4&	82.3&	80.8&	79.6\\
    $\checkmark$ & $\checkmark$ &&	69.1&	83.7&	77.7&	\textcolor{RedOrange}{\textbf{85.2}}&	84.3&	82.5&	80.4\\
    $\checkmark$ & & $\checkmark$ &	67.4&	83.3&	80.5&	83.9&	83.6&	83.7&	80.4\\
    $\checkmark$ & $\checkmark$ & $\checkmark$ &	67.0&	\textcolor{RedOrange}{\textbf{84.2}}&	\textcolor{RedOrange}{\textbf{82.7}}&	84.0&	\textcolor{RedOrange}{\textbf{83.8}}&	\textcolor{RedOrange}{\textbf{84.6}}&	\textcolor{RedOrange}{\textbf{81.0}}\\
    \bottomrule 
    \end{tabular} 
    }
    \label{ablation-lsp} 
\end{table*} 
\subsection{Human Keypoint Detection}\label{sec-human}
\noindent
\textbf{Dataset.} 
\textbf{SURREAL} \cite{varol2017learning} is a synthetic dataset that consists of monocular videos of people with a total of 6M images, which are photo-realistic renderings of people under large variations in shape, texture, viewpoint, and pose. 
Leeds Sports Pose \cite{johnson2010clustered} (\textbf{LSP}) is widely used as the benchmark for human keypoint detection.
LSP contains a total of 2k images of sports persons with annotated human body joint locations gathered from Flickr.

\noindent
\textbf{Implementation Details.}
We use \textbf{SURREAL} as the source domain and \textbf{LSP} as the target domain.
The increasing sequence $M$ in the baby step schedular is set to $\{0.1N,0.2N,0.3N\}$, where $N$ is the total number of training samples, and the number of progressive selection rounds $Q$ is 3.
We report 16 keypoints on the human body parts, \ie, shoulder (Sld), elbow (Elb), wrist, hip, knee, and ankle.

\noindent
\textbf{Results.}
We show the quantitative results in Table \ref{lsp}.
In this task, MAPS does not perform as excellently as the hand keypoint detection benchmark but also obtains a competitive result.
UDAPE achieves the highest accuracy, which benefits from its input-level and output-level alignment modules.
In their input-level alignment module, the style and statistic information of source data is leveraged, which is unavailable in our setting.
Besides, MAPS increases the accuracy of MT by 1.4\%, which is an obvious improvement to validate the effectiveness of the proposed method.
The qualitative results are shown in Fig. \ref{vis-lsp}, we improve the ability of difficult points prediction such as knee and elbow.

\begin{figure*}[t]
\centering
\includegraphics[width=1.8\columnwidth]{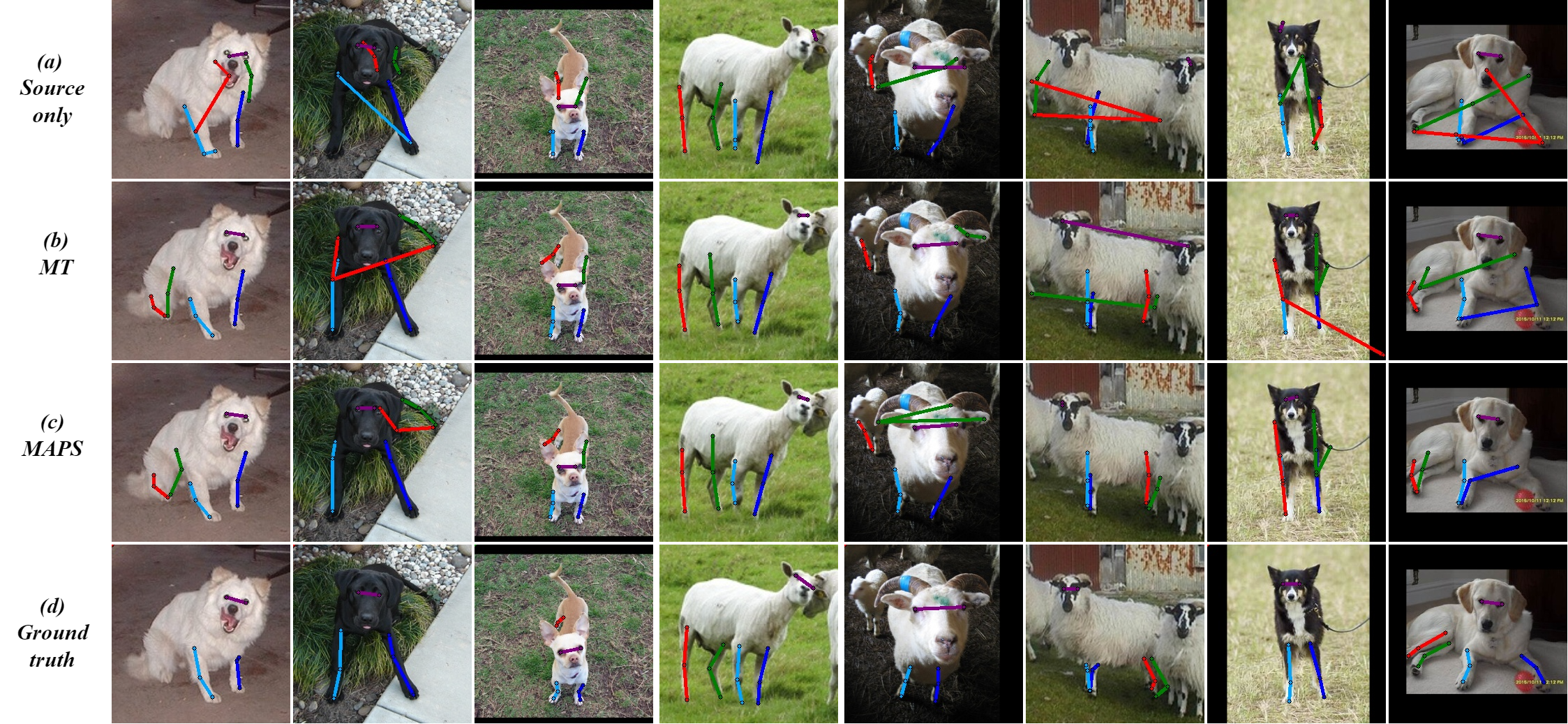}
\centering
\caption{Qualitative results of task \textbf{SynAnimal}$\to$\textbf{AnimalPose}.}
\label{vis-animalpose}
\end{figure*}

\begin{table*}
    \centering 	
    \caption{
    Ablation study on different sample selection strategies on task \textbf{SynAnimal}$\to$\textbf{TigDog}. 
    `Full set': selecting the full target set as easy samples in one round. 
    `One round': selecting part of confident samples in one round. 
    `SPL': selecting part of samples by self-paced learning strategy of MAPS.
    }
    \resizebox{0.75\textwidth}{!}{
    \begin{tabular}{c|cccca|cccca}
    \toprule
    \multirow{2}{*}{Selection} &\multicolumn{5}{c|}{\textbf{SynAnimal}$\to$\textbf{TigDog}: Dog} &\multicolumn{5}{c}{\textbf{SynAnimal}$\to$\textbf{TigDog}: Sheep}\\
    \cmidrule{2-11}
    & Eye&	Hoof&	Knee&	Elb	&Avg. & Eye&	Hoof&	Knee&	Elb&	Avg.\\
    \midrule
    Full set& 58.5&	60.7&	\textcolor{RedOrange}{\textbf{39.4}}&	32.5&	46.2&	69.5&	76.5&	55.6&	\textcolor{RedOrange}{\textbf{46.6}}&	61.0 \\
    One round& 61.9&	\textcolor{RedOrange}{\textbf{61.3}}&	38.6&	\textcolor{RedOrange}{\textbf{34.7}}&	\textcolor{RedOrange}{\textbf{47.3}}&	70.9&	76.0&	56.9&	43.0&	60.4 \\
    SPL& \textcolor{RedOrange}{\textbf{63.2}}&	60.5&	37.3&	34.0&	46.7&	\textcolor{RedOrange}{\textbf{74.2}}&	\textcolor{RedOrange}{\textbf{77.9}}&	\textcolor{RedOrange}{\textbf{58.8}}&	43.3&	\textcolor{RedOrange}{\textbf{62.0}}\\
    \bottomrule 
    \end{tabular}
    }
    \label{ablation-selection-animal} 
\end{table*} 

\begin{table*}
    \centering 	
    \caption{
    Ablation study on different sample selection strategies on task \textbf{SURREAL}$\to$\textbf{LSP}. 
    }
    \resizebox{0.55\textwidth}{!}{     	 
    \begin{tabular}{c|cccccca}
    \toprule
    \multirow{2}{*}{Selection} & \multicolumn{7}{c}{\textbf{SURREAL}$\to$\textbf{LSP}}\\
    \cmidrule{2-8}
    &Sld& Elb &Wrist& Hip& Knee& Ankle& Avg. \\
    \midrule
    Full set&	54.3&	67.2&	65.6&	73.8&	73.4&	73.0&	67.9\\
    One round& 	\textcolor{RedOrange}{\textbf{65.3}}&	81.1&	79.3&	83.8&	84.0&	84.1&	79.6\\
    SPL& 64.4&	\textcolor{RedOrange}{\textbf{84.3}}&	\textcolor{RedOrange}{\textbf{81.5}}&	\textcolor{RedOrange}{\textbf{84.4}}&	\textcolor{RedOrange}{\textbf{84.5}}&	\textcolor{RedOrange}{\textbf{84.5}}&	\textcolor{RedOrange}{\textbf{80.6}}\\
    \bottomrule 
    \end{tabular} 
    }
    \label{ablation-selection-lsp} 
\end{table*} 

\begin{figure*}
    \centering   	 
    \begin{tabular}{ccc}     		 
    \includegraphics[scale=0.33]{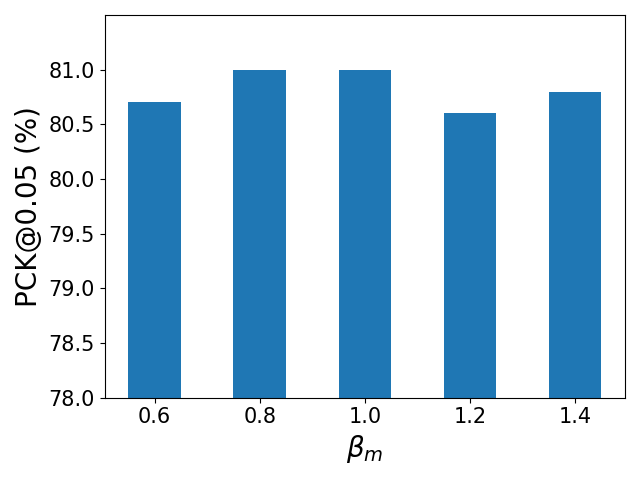} &\includegraphics[scale=0.33]{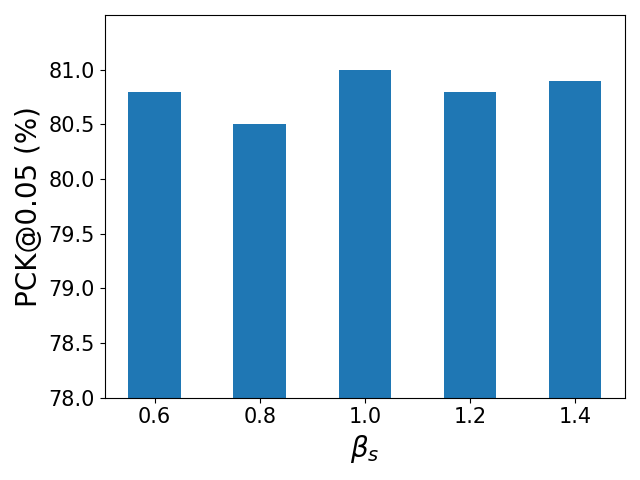}&
    \includegraphics[scale=0.33]{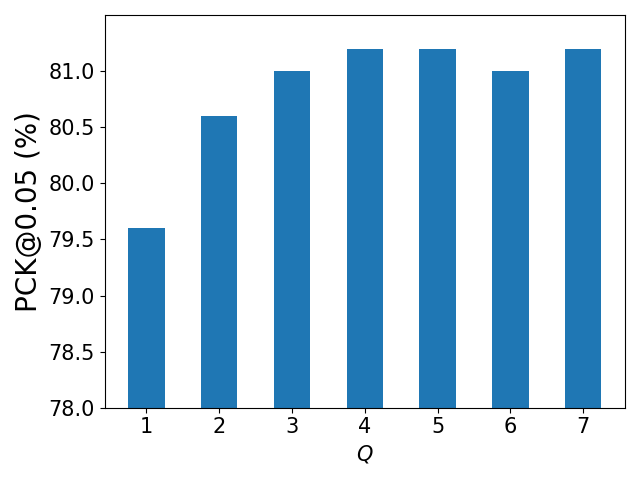}\\
    (a) Influence of the weight $ \beta_{m}$. &(b) Influence of the weight $ \beta_{s}$.& (c) Influence of the rounds $Q$.
    
    \end{tabular} 
    \caption{Analysis of the weights $\beta_m$ and $\beta_s$, and the number of progressive selection rounds $Q$ on task \textbf{SURREAL}$\to$\textbf{LSP}.}
    \label{weights} 
\end{figure*}

\subsection{Animal Keypoint Detection}\label{sec-animal}
\noindent
\textbf{Dataset.}
Synthetic Animal Dataset \cite{mu2020learning} (\textbf{SynAnimal}) is a synthetic pose dataset rendered from CAD models, containing 5 different animals including the horse, tiger, sheep, hound, and elephant. 
Each class contains 10k images. 
\textbf{AnimalPose} dataset \cite{Cao_2019_ICCV} contains 6.1k images in the wild from 5 animals, \ie, dog, cat, cow, sheep, and horse.
\textbf{TigDog} dataset \cite{del2015articulated} is a large-scale animal keypoint detection dataset, which provides 30k images from real-world videos of horse and tigers.

\noindent
\textbf{Implementation Details.}
We conduct two experiments in this section, with \textbf{SynAnimal} as the source domain in both experiments, \textbf{AnimalPose} and \textbf{TigDog} as the target domain, respectively. 
In \textbf{SynAnimal}$\to$ \textbf{AnimalPose}, the increasing sequence $M$ is set to $\{0.3N,0.4N,0.5N\}$, and the number of progressive selection rounds $Q$ is set to 3.
We report the detection results of 14 keypoints on the dog and sheep body parts including the eye, hoof, knee, and elbow (Elb).
In \textbf{SynAnimal}$\to$ \textbf{TigDog}, $M$ is set to $\{0.1N,0.15N,0.2N\}$, and $Q$ is also set to 3.
We present the detection results of 14 keypoints on the horse and tiger body parts including the eye, chin, shoulder (Sld), hip, elbow (Elb), knee, and hoof.
We run these experiments with the same augmentation in UPADE.

\noindent
\textbf{Results}
The results of \textbf{SynAnimal}$\to$\textbf{AnimalPose} are shown in Table \ref{animalpose}.
MAPS achieves the best results among these methods on both dog and sheep keypoint detection and improves the baseline method MT by a large margin (2\%).
\textbf{AnimalPose} is a small dataset, which makes the model easily overfitting to outliers, and our self-mixup loss favors the predictions to vary linearly with the input to improve the robustness, therefore performing well on such a small dataset.
The qualitative results are shown in Fig. \ref{vis-animalpose}.
The results of \textbf{SynAnimal}$\to$ \textbf{TigDog} are shown in Table \ref{tigdog}.
In this task, MAPS still has room for improvement.
The reason why MAPS performs not well is probably due to the fact that the domain gap between \textbf{SynAnimal} and \textbf{TigDog} is primarily reflected in image style.
UDAPE \cite{kim2022unified} directly solves the style alignment by introducing a pretrained style transfer model.
UDA-Animal \cite{li2021synthetic}, as a typical animal keypoint detection method, narrows the domain gap through adversarial training, which relies on the statistical information in source images.
The lack of the style information of the source samples suppresses the performance of MAPS on this task, but MAPS still outperforms the MT, which demonstrates our method works in large-scale scenarios.

\subsection{Ablation Study}\label{sec-ablation}
To explore the effectiveness of the proposed strategies in different scenarios, 
For fairness, we conduct the experiments on two benchmarks, \ie, \textbf{SynAnimal}$\to$\textbf{AnimalPose} and \textbf{SURREAL}$\to$\textbf{LSP}.

\noindent
\textbf{Effectiveness of loss functions.}
We first study the effectiveness of the three loss functions and the interactions among them.
As shown in Table \ref{ablation-animal} and \ref{ablation-lsp}, the teacher-student consistency learning framework with the mean-teacher loss $\mathcal{L}_{mt}$ constructs a strong baseline method.
The self-mixup loss $\mathcal{L}_m$ and the self-paced loss $\mathcal{L}_s$ improve the results of baseline by about 1\% on the animal benchmark and 0.8\% on the human benchmark, respectively.
With both $\mathcal{L}_m$ and $\mathcal{L}_s$, the accuracy of the baseline is improved by about 2\% on the animal benchmark and 1.6\% on the human benchmark, demonstrating that both two proposed loss functions are practical, and the interactions among three loss functions are positive and useful.

\noindent
\textbf{Effectiveness of sample selection strategies.}
We conduct an ablation study on different sample selection methods.
The results are presented in Table \ref{ablation-selection-animal} and \ref{ablation-selection-lsp}.
In the tables,
`Full set' denotes selecting the full target set as easy samples in one round.
`One round' denotes selecting part of confident samples in one round.
This can be seen as a special case of progressive selection, where $Q=1$.
`SPL' denotes the progressive selection strategy in MAPS, \ie, selecting part of samples by self-paced learning strategy.
On both two tasks, the 'SPL' row obtains the highest accuracy, which demonstrates that selecting the part of confident samples in one round filters out most of the noisy pseudo labels, and the performance is improved.
On task \textbf{SynAnimal}$\to$\textbf{AnimalPose}, selecting the full target set as easy samples is better than selecting part of confident samples in one round.
Although the full target set has more noisy labels, it also contains a larger amount of data and more information.
The benefits of the large data and information can compensate for the negative effects of noise.
On task \textbf{SURREAL}$\to$\textbf{LSP}, the opposite is true.
The negative effect of noise from using the full dataset is severe, and selecting part of confident samples alleviates this issue.
In fact, this is a trade-off problem between noise and information, our progressive selection strategy improves the pseudo-label quality by leveraging the dynamic capacity of the model to achieve the best results.

\subsection{Further Analysis}\label{sec-further}

\noindent
\textbf{Sensitivity of the weights.}
We conduct sensitivity analysis on two major hyper-parameters, \ie, weights of loss functions, in our framework.
We empirically choose value 1.0 for both two hyper-parameters.
In Fig. \ref{weights} (a), the accuracy around $\beta_m=1.0$ is stable at 80.5\% to 81.0\%.
In Fig. \ref{weights} (b), the accuracy around $\beta_s=1.0$ is also stable, floating around 0.5\%.
In summary, our weights of loss functions are not sensitive.

\noindent
\textbf{Sensitivity of the number of selection rounds.}
As shown in Fig. \ref{weights} (c), along with the increasing of progressive learning rounds $Q$, the accuracy increases from around 79.5\% to around 81.5\%.
This is reasonable, because the smaller the rounds $Q$, the greater the noise of the selected easy samples, leading to worse performance.
When it grows to $Q=3$, the accuracy gradually stabilizes and does not bring greater gains as $Q$ grows.
Therefore, we choose $Q=3$ in our experiments.

\section{Conclusion}
This paper considers a realistic setting called source-free domain adaptive keypoint detection, where only the well-trained source model is provided to the target domain.
We first construct a simple baseline method based on the mean-teacher framework.
Then we propose a new approach termed Mixup Augmentation and Progressive Selection (MAPS) built on this.
The mixup augmentation regularizes the model to favor simple linear behavior in-between target samples thereby improving the robustness.
The progressive selection strategy leverages the dynamic capacity of the current model to explore reliable pseudo labels. 
These two strategies respectively consider the convex behavior among samples and the quality of the pseudo label per sample, making them complement each other.
Experiments verify that MAPS achieves competitive and even state-of-the-art performance in comparison to previous non-source-free counterparts.

\newpage 
\bibliographystyle{IEEEtran} 
\bibliography{ref}

\end{document}